\documentclass{article}

     \PassOptionsToPackage{numbers, compress}{natbib}

\usepackage[preprint]{neurips_2020}

\usepackage[utf8]{inputenc} 
\usepackage[T1]{fontenc}    
\usepackage{hyperref}       
\usepackage{url}            
\usepackage{booktabs}       
\usepackage{amsfonts}       
\usepackage{nicefrac}       
\usepackage{microtype}      
\usepackage{graphicx}
\usepackage{multirow}
\usepackage{paralist}
\usepackage{bm}
\usepackage{amsmath}
\newcommand{\specialcell}[2][l]{%
  \begin{tabular}[#1]{@{}l@{}}#2\end{tabular}}

\title{Generalized Zero-Shot Domain Adaptation via Coupled Conditional Variational Autoencoders}

\author{%
  Qian Wang\thanks{Use footnote for providing further information
    about author (webpage, alternative address)---\emph{not} for acknowledging
    funding agencies.} \\
  Department of Computer Science\\
  Durham University\\
  Durham, United Kingdom \\
  \texttt{qian.wang173@hotmail.com} \\
   \And
   Toby P. Breckon \\
  Department of \{Computer Science | Engineering\}\\
  Durham University\\
  Durham, United Kingdom \\
   \texttt{toby.breckon@durham.ac.uk} \\
}

\begin{document}

\maketitle

\begin{abstract}
Domain adaptation approaches aim to exploit useful information from the source domain where supervised learning examples are easier to obtain to address a learning problem in the target domain where there is no or limited availability of such examples. In classification problems, domain adaptation has been studied under varying supervised, unsupervised and semi-supervised conditions. However, a common situation when the labelled samples are available for a subset of target domain classes has been overlooked. In this paper, we formulate this particular domain adaptation problem within a generalized zero-shot learning framework by treating the labelled source domain samples as semantic representations for zero-shot learning. For this particular problem, neither conventional domain adaptation approaches nor zero-shot learning algorithms directly apply. To address this generalized zero-shot domain adaptation problem, we present a novel Coupled Conditional Variational  Autoencoder (CCVAE) which can generate synthetic target domain features for unseen classes from their source domain counterparts. Extensive experiments have been conducted on three domain adaptation datasets including a bespoke X-ray security checkpoint dataset to simulate a real-world application in aviation security. The results demonstrate the effectiveness of our proposed approach both against established benchmarks and in terms of real-world applicability.
\end{abstract}

\section{Introduction}
The success of deep learning in the recent decade relies on the availability of abundant annotated data for training. In real-world applications, the acquisition of sufficient training data can be difficult or even impossible. One technique to address the training data sparsity issue is transfer learning which aims to explore and transfer knowledge learned from the source domain to the target domain. There are usually more annotated data in the source domain than those in the target domain within which the task to solve resides. \textit{Zero-shot learning} \cite{wang2017zero,wang2017alternative,xian2019f} and \textit{domain adaptation} \cite{wang2019unifying,wang2020unsupervised} are two well-formulated transfer learning problems that have attracted much attention in the recent decade.

Traditional supervised learning methods have the limitation in that they can only recognize \textit{seen classes} (observed) for which labelled samples are available during training. In contrast, zero-shot learning aims to recognize samples from not only seen classes but also novel unseen classes (unobserved) for which no training samples are available during training \cite{wang2017zero}. To this end, side information of both seen and unseen classes from a \textit{source domain} (in contrast to the \textit{target domain} where the recognition task resides) is needed to model the between-class relations. In zero-shot visual recognition, class-level semantic attributes or word vectors are usually adopted as the side information in the source domain (i.e. semantic representation space) whilst the image classification task is addressed in the target domain (i.e. visual representation space) \cite{wang2017alternative}. 

Domain adaptation is a technique aiming to mitigate the distribution discrepancy between source and target domains so that the knowledge learned from annotated source domain samples can be applied in the target domain \cite{wang2019unifying,wang2020unsupervised}. In this sense, zero-shot learning can be seen as a specific type of domain adaptation problem where the source domain provides per-class samples while the target domain provides training samples only for certain seen classes. In practice, however, existing domain adaptation approaches to unsupervised, supervised or semi-supervised domain adaptation in literature are not readily applicable to the current zero-shot learning problem setting. The underlying reason for this is twofold. On one hand, current zero-shot learning problems are constrained to have only class-level semantic representations in the source domain hence it is extremely difficult to learn sufficient knowledge for seen to unseen classes transfer \cite{wang2017alternative,wang2019unifying}. On the other hand, most domain adaptation approaches take advantage of a certain number of target domain samples (either labelled or not) during training which hinders its application in scenarios where such target domain samples are not available for all classes. To fill in this gap, zero-shot domain adaptation problems have been studied in \cite{peng2018zero,wang2019conditional,ishii2020partially} assuming that there are plenty of labelled samples in the source domain for all concerned classes, whilst labelled samples are available for only a subset of seen classes in the target domain. However, these studies restrict the capability of recognition in the unseen class space. As agreed in the zero-shot learning literature, generalized zero-shot learning, in which the recognition of both seen and unseen classes in the target domain is required, is more practically useful. In the same spirit, we take one step further in this paper to address a novel Generalized Zero-Shot Domain Adaptation (GZSDA) problem arising from many real-world applications.

In the scenario of image classification,  the relations of the novel generalized zero-shot domain adaptation problem with traditional zero-shot learning and unsupervised/supervised domain adaptation problems are illustrated in Table \ref{table:definition}.
The data imbalance across domains and classes poses more challenges to the GZSDA problem hence existing approaches to either zero-shot learning or domain adaptation tend to bias to either seen classes or the source domain.

To attack the data imbalance issue in the GZSDA problem, we present a novel Coupled Conditional Variational Autoencoder (CCVAE) solution by generating unseen data in the target domain to re-balance the training data. Specifically, the proposed CCVAE is able to transform source domain samples into associated projections within the target domain without loss of class information and vice versa. As a result, target domain samples of unseen classes can be generated from the corresponding source domain samples and used to train a classifier for all classes in a traditional supervised learning manner. The CCVAE works in the feature space rather than the image pixel space to reduce the complexity and challenge of image generation since the goal of GZSDA is image classification rather than image generation. 
Following this outline, the contributions of this paper can be summarized as follows:
\begin{compactitem}
\item[--] a novel Generalized Zero-Shot Domain Adaptation (GZSDA) problem is formulated and studied for the first time extending the prior definitions of ZSDA problems in this domain from \cite{peng2018zero,wang2019conditional}; such a GZSDA problem is more realistic and also more challenging to solve.
\item[--] a novel Coupled Conditional Variational Autoencoder (CCVAE) model is proposed to address the GZSDA problem extending and outperforming the prior work of \cite{wang2019unifying}; the proposed CCVAE integrates the benefits of feature transformation and feature generation in one framework.
\item[--] a new multi-domain dataset arising from real-world applications is collected, annotated and released for domain adaptation research; it comprises of cross spectral image domains (i.e. dual-energy colour-mapped X-ray and regular colour photograph) which are not present in other datasets.
\item[--] extended experimentation is performed on two benchmark datasets in addition to a bespoke X-ray security checkpoint dataset to validate the effectiveness of the proposed CCVAE in GZSDA problems both against established benchamrks and in terms of real-world applicability; and also its superiority to a variety of contemporary methods in the field.
\end{compactitem}

\section{Related Work}\label{sec:related}
We review closely related work to our study from the perspective of zero-shot learning, domain adaptation and zero-shot domain adaptation and summarize the relationship to existing research topics and approaches in Table \ref{table:definition}.

\textbf{Domain Adaptation}\\
Existing domain adaptation approaches \cite{wang2019unifying,wang2020unsupervised} try to align the marginal distributions across source and target domains \cite{wang2020unsupervised} or to learn domain-invariant representations \cite{pei2018multi} so that labeling information available in the source domain can be explored to guide the learning of a classifier in the target domain or a latent common space. 
Fine-grained class-wise adaptation across domains has been employed by promoting the alignment of conditional distributions as an additional constraint \cite{long2018conditional}. 
For unsupervised domain adaptation, this can be implemented by pseudo-labeling \cite{chen2019progressive,wang2020unsupervised} given the access to unlabelled target domain samples for all classes. However, in the scenario of zero-shot domain adaptation, class-wise adaptation forms the primary challenge that we address in this work due to the lack of samples for unseen classes in the target domain regardless of labelled or unlabelled. 

\begin{table*}[!thp]
    \centering
    {
        \centering
        \caption[]{A comparison of generalized zero-shot domain adaptation with other related research topics.}
        \label{table:definition}
        \resizebox{0.9\textwidth}{!}{%
            \begin{tabular}{l|l|l|l}
                \hline
                \multirow{2}{*}{Research Problem} & \multicolumn{2}{c|}{Training} &Testing\\
                \cline{2-4}
                & Source & Target & Target \\ \hline
                \specialcell{Unsupervised Domain Adaptation\\ (UDA, \cite{wang2019unifying, wang2020unsupervised})} & labelled samples for all classes & \specialcell{unlabelled samples for all classes \\ (same as testing)} &  all classes \\ \hline
                \specialcell{Supervised Domain Adaptation\\ (SDA, \cite{motiian2017unified})} & labelled samples for all classes & \specialcell{labelled samples for all classes \\ (a small number)} & all classes\\ \hline
                \specialcell{Zero-Shot Learning (ZSL, \cite{wang2017zero})} & per-class representations for all classes & labelled samples for seen classes & unseen classes\\ \hline
                \specialcell{Generalized ZSL (GZSL, \cite{mishra2018generative})} & per-class representations for all classes & labelled samples for seen classes & all classes \\\hline
                \specialcell{Zero-Shot Domain Adaptation\\ (ZSDA, \cite{peng2018zero,wang2019conditional,ishii2019zero})} & \specialcell{unlabelled samples for seen classes and \\ labelled samples for unseen classes} & unlabelled samples for seen classes & unseen classes \\ \hline
                \textbf{\specialcell{Generalized Zero-Shot Domain\\ Adaptation (GZSDA, this paper)}} & labelled samples for all classes & labelled samples for seen classes & all classes \\
                \hline
            \end{tabular}%
        }%
    }
\end{table*}

\textbf{Zero-Shot Learning}\\
The most popular approaches to zero-shot learning are based on generative model such as Generative Adversarial Networks (GAN) \cite{xian2019f} and Varitional Autoencoders (VAE) \cite{mishra2018generative}. The generative models are trained to generate image features for specific classes given the corresponding class-level semantic representations (i.e. attributes or word vectors). Subsequently, a classifier can be trained using the combined real and generated data covering both seen and unseen classes.
On intrinsic limitation of ZSL is that the class-level attributes or word vectors in the source domain restrict the capability of representing the intra-class variability. 
Alternatively, the class-level semantic representations in zero-shot learning can be replaced by more informative labelled samples in a source domain where such labelled samples are easy to collect and annotate. This leads to the very novel zero-shot learning problem we focus on in this paper. Existing zero-shot learning methods \cite{xian2019f,mishra2018generative} can not be directly applied to this problem since the source domain information appears in a different modality, whilst the ideas of generating synthetic image features for unseen classes will be employed and extended in our approach (Section \ref{sec:method}).

\textbf{Zero-Shot Domain Adaptation}\\
Very limited prior work has addressed zero-shot domain adaptation problems. 
Yang et al. \cite{yang2015zero} attempted to address the issue where multiple source domains and the target domain are determined by a vector of continuous variables. Here there is no data available for the target domain but the corresponding control variables are known as prior knowledge. The transfer learning across source and target domains can be explicitly modelled by such control variables. 
Similar assumptions are made by \cite{ishii2019zero} which assumes that prior knowledge of attribute information exists (e.g., time, angle, gender, age, etc.) characterizing the difference between source and target domains. 
In contrast, we aim to address a more generic problem without the need of these control variables relating source and target domains.
The problem we try to address in this work is more similar to that in \cite{peng2018zero,wang2019conditional,ishii2020partially} which instead restrict the recognition to unseen classes. Moreover, paired task-irrelevant data (i.e. seen class data in this context) from source and target domains are required during training in \cite{peng2018zero} and \cite{wang2019conditional} and such correspondences may not exist in the labelled samples from seen classes in most real cases. We lift these restrictions and focus on the GZSDA problem without the need of either control variables or paired training samples.

\section{Method} \label{sec:method}
In this section, we first describe the problem settings of Generalized Zero-Shot Domain Adaptation and subsequently our proposed solution to this problem. 

\begin{figure}
    \centering
    {\includegraphics[width=0.8\textwidth]{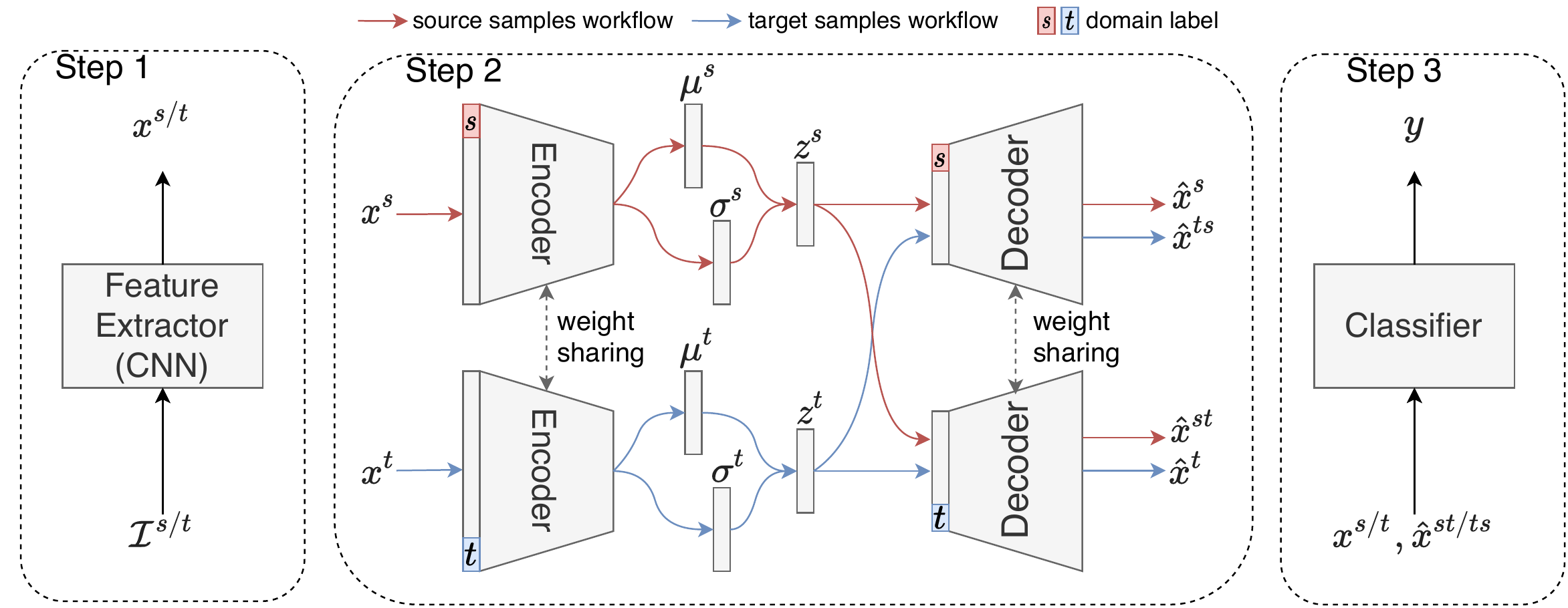}}
    {\caption{Our proposed Coupled Conditional Variational Autoencoder (CCVAE) framework.}
    \label{fig:ccvae_framework}}
\end{figure}

Our proposed framework consists of three steps as illustrated Figure \ref{fig:ccvae_framework} (full details in Section \ref{sec:ccvae}). Step one trains a feature extractor using all the labelled training data from both domains. In the second step, a Coupled Conditional Variational Autoencoder is trained using image features extracted in step one and will be used to generate synthetic features in the target domain. With the combination of these synthetic features and features extracted from real trainig images, a classifier is trained and used for image classification in the target domain.

\subsection{Problem Formulation of Generalized Zero-Shot Domain Adaptation}

Given a labelled dataset $\mathcal{D}^s = \{(x^s_i,y^s_i)\}, i = 1,2,...,n^s$ from the source domain $\mathcal{S}$, $x^s_i$ represents the $i$-th training sample (e.g., an image in our case) in the source domain, and $y^s_i \in \mathcal{Y} = \{1,2,...,C\}$ denotes the corresponding label, and $C$ is the number classes. In the target domain, a labelled dataset $\mathcal{D}^{t}=\{(x_i^{t},y_i^{t})\},i=1,2,...,n^{t}$ from the target domain $\mathcal{T}$. $x_i^{t}$ and $y_i^{t} \in \mathcal{Y}^{seen}$ are the $i$-th labelled sample and its label respectively. Note that $\mathcal{Y}^{seen} \subset \mathcal{Y}$, that is, labelled samples are available for only a subset of classes in the target domain. The label space $\mathcal{Y} = \mathcal{Y}^{seen} \cup \mathcal{Y}^{unseen}$ is shared by source and target domains. The task is to classify any given new instance $x$ from the target domain by learning an inference model $y=f(x) \in \mathcal{Y}$.

\subsection{Feature Extraction}
The key of our approach to the GZSDA problem is the generation of synthetic data for unseen classes in the target domain. Given the challenge of image generation in the pixel space \cite{xian2019f}, we choose to generate image features since the ultimate goal is image classification rather than image generation. To this end, we extract image features in the first step. As shown in Figure \ref{fig:ccvae_framework}, a shared deep Convolutional Neural Network (CNN) model is employed to extract features for images from both source and target domains. We use ResNet50 \cite{he2016deep} pre-trained on the ImageNet \cite{deng2009imagenet} as the feature extractor for object images in our experiments and AlexNet \cite{krizhevsky2012imagenet} trained from scratch using $\mathcal{D}^s$ and $\mathcal{D}^t$ for digits data. We will use $\bm{x}$ and $\bm{\tilde{x}}$ to denote the real and synthetically generated image features in the following sections.

\subsection{Coupled Conditional Variational Autoencoder}
\label{sec:ccvae}
\textbf{Variational Autoencoder}\\
The Variational Autoencoder \cite{kingma2013auto} encodes an input feature $\bm{x}$ into a distribution $p_{\theta}(\bm{z})$ (approximated by $q_{\Phi}(\bm{z}|\bm{x})$) from which the latent encoding vector $\bm{z}$ can be sampled and subsequently fed into the decoder to reconstruct the input feature $\tilde{\bm{x}}$.  The decoder can be parameterized by $p_{\theta}(\bm{x}|\bm{z})$. According to \cite{kingma2013auto}, the objective function for the VAE can be written as follows:
\begin{equation}\label{eq:vaeLoss}
\mathcal{J}_{VAE}(\Phi,\theta;\bm{x}) = -D_{KL}(q_{\Phi}(\bm{z}|\bm{x}) || p_{\theta}(\bm{z})) + \mathbb{E}_{q_{\Phi}(\bm{z}|\bm{x})} [\log p_{\theta}(\bm{x}|\bm{z}) ]
\end{equation}
where  $D_{KL}(p||q)$ is the Kullback-Leibler (KL) divergence between two distributions $p$ and $q$. The VAE is trained by maximizing $\mathcal{L}(\Phi,\theta;\bm{x})$ which can be interpreted as minimizing the reconstruction error and the KL divergence.

\textbf{Conditional VAE}\\
Conditional VAE (CVAE) was first proposed in \cite{sohn2015learning}. It allows for modeling multiple modes (e.g., classes) in conditional distribution of the target variable (e.g., reconstructed input $\bm{\tilde{x}}$) given input $\bm{x}$ and the condition $c$. The objective function of CVAE can be adapted from Eq.(\ref{eq:vaeLoss}) as follows:
\begin{equation}\label{eq:cvaeLoss}
\mathcal{J}_{CVAE}(\Phi,\theta;\bm{x},c) = -D_{KL}(q_{\Phi}(\bm{z}|\bm{x},c) || p_{\theta}(\bm{z})) + \mathbb{E}_{q_{\Phi}(\bm{z}|\bm{x},c)} [\log p_{\theta}(\bm{x}|\bm{z},c) ]
\end{equation}
In existing CVAE models \cite{mishra2018generative,yan2016attribute2image}, both the encoder $q_{\Phi}(\bm{z}|\bm{x},c)$ and the decoder $p_{\theta}(\bm{x}|\bm{z},c)$ are conditioned on the class information (e.g., class-wise attributes). In ZSL problem \cite{mishra2018generative}, the CVAE is trained using target domain data in the condition of class-wise attributes from the source domain. Each training sample in the target domain has its corresponding attribute vector as the condition in the CVAE model. However, in our GZSDA problem information from the source domain is represented by labelled samples rather than class-level representations. In addition, the cross-domain correspondence in the sample level is unavailable. The traditional CVAE is not applicable to this problem.


\textbf{Coupled Conditional VAE}\\
The challenge of GZSDA problem originates from the missing labelled samples for unseen classes in the target domain. We attempt to learn a generative model based on CVAE to generate synthetic features for unseen classes in the target domain. The generated features are required to be both class discriminative and domain discriminative. To these ends, the decoder $p(\bm{x}|\bm{z},c)$ in the CVAE is conditioned on domain labels to generate domain discriminative features whilst the latent codes $\bm{z}$  need to be class discriminative to generate class discriminative features.

The proposed CCVAE is illustrated in Figure \ref{fig:ccvae_framework} (step 2). It is composed of a pair of CVAE for the source and target domains respectively. In our work, we model both the encoders and decoders using fully-connected neural networks. We force two CVAE to have identical architectures with shared weights. As a result, the model degrades into one coupled CVAE trained on both source and target domain data. 

During training, the encoder takes the concatenation of a feature vector $\bm{x^s}/\bm{x^t}$ from the source/target domain and its corresponding domain label $c(\bm{x})= s/t$ (represented by a one-hot 2-dimensional vector) as the input to estimate the latent code distribution $q(\bm{z}|\bm{x},c) = \mathcal{N}(\mu_{\bm{x}},\Sigma_{\bm{x}})$. Subsequently, a latent code $\bm{z}$ is sampled from $\mathcal{N}(\mu_{\bm{x}},\Sigma_{\bm{x}})$ and fed into the decoder with the same domain label $s/t$ as the condition to reconstruct the input as $\bm{\tilde{x}^s}/\bm{\tilde{x}^t}$. On the other hand, the sampled latent code $\bm{z}$ can also be decoded with the condition of the other domain label $t/s$ to generate the synthetic feature in a different domain as $\bm{\tilde{x}^{st}}/\bm{\tilde{x}^{ts}}$.

The model is trained by feeding paired source and target domain samples $\{\bm{x}^s,\bm{x}^t\}$ randomly selected from the same class. The loss function to minimize can be formulated as:
\begin{equation}\label{eq:ccvaeLoss}
\begin{aligned}
\mathcal{L}_{CCVAE}(\Phi,\theta;\bm{x^s},\bm{x^t}) = & (\mathcal{L}_{recon} \big(\bm{x^s},\bm{\tilde{x}^s}) + \mathcal{L}_{recon} (\bm{x^t},\bm{\tilde{x}^t})\big) \\ 
&+  \big(\mathcal{L}_{cross\_recon} (\bm{x^s},\bm{\tilde{x}^{ts}}) + \mathcal{L}_{cross\_recon} (\bm{x^t},\bm{\tilde{x}^{st}})\big)\\
& +  \lambda D_{KL}\big(\mathcal{N}(\mu_{\bm{x}},\Sigma_{\bm{x}}) || \mathcal{N}(\bm{0},\bm{I})\big) 
\end{aligned}
\end{equation}
The first terms measure the reconstruction errors for both source and target domain samples. The second terms measure the cross-domain reconstruction errors. Although the samples in the pair of $\{\bm{x^s},\bm{\tilde{x}^{ts}}\}$ or $\{\bm{x^t},\bm{\tilde{x}^{st}}\}$ are from the same class, they are not necessarily two views of the same image. To reduce the cross-domain reconstruction errors, the encoder has to preserve class information in the latent code space. As a result, the use of cross-domain reconstruction loss $\mathcal{L}_{cross\_recon}$ facilitate the model to generate class discriminative features across domains. 
For those $\bm{x}^s$ belonging to unseen classes, there is no valid target domain samples $\bm{x}^t$ from the same class. We use dummy features in practice and exclude the loss terms involving these dummy features. 
The third term aims to reduce the KL divergence between the distributions of the latent code and a normal distribution. It serves as a regularization term in the same way as in the VAE model. 
$\lambda$ is a hyper-parameter balancing the KL divergence and reconstruction errors. We use L2 loss for both $\mathcal{L}_{recon}$ and $\mathcal{L}_{cross\_recon}$. The effectiveness of different terms in Eq.(\ref{eq:ccvaeLoss}) will be further investigated and discussed in Section \ref{sec:experiments}.

\subsection{Target Image Classification} \label{sec:step3}
Once the CCVAE is trained, we can use it to generate synthetic features by the cross-domain reconstruction pipelines. Specifically, given a feature vector $\bm{x}^s$ (or $\bm{x}^t$), the model can generate $\bm{\tilde{x}^{st}}$ (or $\bm{\tilde{x}^{ts}}$) which should have the same class label as the input. In this way, we are able to generate synthetic features for unseen classes in the target domain with the source domain samples. We use real data $\mathcal{D}^s$ and $\mathcal{D}^t$ together with synthetically generated features from them to train a unified neural network classifier for all classes and both domains. The classifier is then used to classify test images.

\section{Experiments and Results}\label{sec:experiments}
As the first attempt to address the GZSDA problem, we present a benchmark on GZSDA with extensive experiments on three datasets. We compare our proposed CCVAE with baseline methods and state-of-the-art methods for zero-shot learning \cite{wang2017zero,mishra2018generative,wang2019unifying} which have been adapted to the GZSDA problem. 

\subsection{Dataset}
\begin{figure}
    \centering
    {\includegraphics[width=0.8\textwidth]{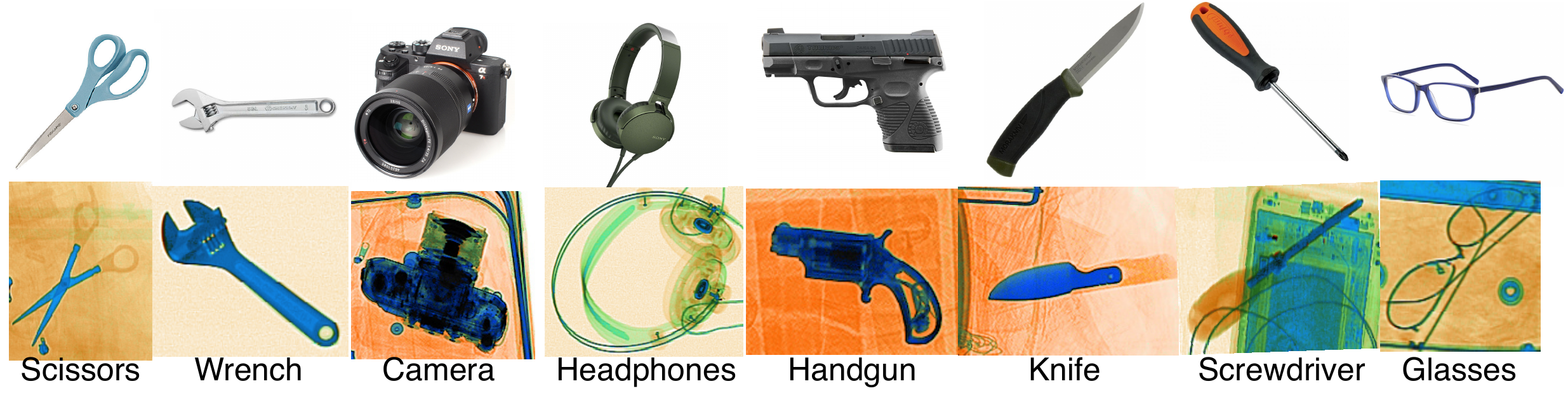}}
    {\caption{Sample images from the BaggageXray-20 dataset (upper: regular; bottom: X-ray).}
    \label{fig:xray}}
\end{figure}

\textbf{BaggageXray-20}\\
This dataset is collected for the purpose of automatic object recognition in aviation security baggage screening \footnote{The dataset will be publicly and openly realeased including annotations, at permanently hosted DOI.}. Compared with the prevalence of regular RGB images, X-ray images are rarely available and have significantly different appearances as shown in Figure \ref{fig:xray}. The dataset consists of two domains (i.e. colour mapped X-ray images and regular visible band photographic images) covering 20 object classes. There are 4620 and  3444 images in the X-ray and regular domains, respectively. We use ResNet101 pretrained on the ImageNet to extract 2048-dim features for images from both domains. To simulate the GZSDA problem setting, the 20 classes were randomly split into two subsets: 10 classes as the seen classes and the rest 10 classes as the unseen classes. Five random seen/unseen class splits were used in our experiments to get statistics of the experimental results. Two domain adaptation tasks (i.e. regular $\to$ X-ray and X-ray $\to$ regular) were employed in the experiments. When a domain serves as the source domain, all images are used to form the labelled source dataset $\mathcal{D}^s$ whilst for the domain serving as the target domain we randomly reserve 50\% of the images from each class for testing and the rest 50\% from seen classes form the labelled target dataset $\mathcal{D}^t$.

\textbf{Office-Home}\\
Office-Home \cite{venkateswara2017deep} is a dataset commonly used for domain adaptation. It consists of four domains: artistic images (Art), Clipart, Product images and Real-World images. There are 65 object classes in each domain with a total number of 15,588 images. We extract ResNet50 features in our experiments and divide these 65 classes into 35 seen classes and 30 unseen classes randomly and generated 5 different splits for calculating the statistics of results. Given four domains, there could be 12 different domain adaptation tasks among which we report results with \textit{ClipArt} as the source domain in this paper. The training and test dataset creation strategy is identical to BaggageXray dataset.

\textbf{MNIST/Fashion-MNIST/EMNIST (\textit{X}MNIST)}\\
We follow previous works on ZSDA \cite{peng2018zero,wang2019conditional} and conducted experiments using MNIST \cite{lecun1998gradient}, Fashion-MNIST (FMNIST) \cite{xiao2017fashion} and EMNIST \cite{cohen2017emnist} (denoted collectively as \textit{X}MNIST). These three datasets contain 10, 10 and 26 classes respectively. In MNIST and FMNIST datasets, there are 6000 and 1000 images in the training and test subsets for each class, respectively. In the EMNIST dataset, the number of training and test images per class are 4800 and 800 respectively.
All these three datasets contain gray images of the same size of 28$\times$28. We consider these gray images as in the \textit{Gray} domain from which we create another two domains \textit{Color} and \textit{Negative}. The \textit{Color} domain images were created using the method proposed in \cite{ganin2015unsupervised}. Specifically, for a given image $I$, a random patch $P$ of the same size was cropped from a color image in BSDS500 \cite{arbelaez2010contour} and the color version of $I$ is created by $I_c = |I-P|$ for all channels. The \textit{Negative} domain images are obtained by $I_n = 255 - I$.
There are 6 different combinations of 3 domains to form 6 domain adaptation tasks among which we report the representative results of \textit{Gray} $\to$ \textit{Color}, \textit{Color} $\to$ \textit{Gray} and \textit{Negative} $\to$ \textit{Color}. In each domain adaptation task, we choose any two datasets as the seen and unseen classes respectively. As a result, for each domain adaptation task there can be 6 sub-tasks with different combinations of seen and unseen datasets. AlexNet \cite{krizhevsky2012imagenet} was trained from scratch using the training data $\mathcal{D}^s$ and $\mathcal{D}^t$ to extract features for a specific adaptation task.

\subsection{Implementation Details}
The proposed method was implemented in PyTorch\footnote{The code will be released.} \cite{paszke2019pytorch}. Both the encoder and decoder were three-layer fully-connected neural networks. For BaggageXray-20 and Office-Home datasets, the VAE share the same architecture of $2048-512-64-64-512-2048$ where 64 is the dimension of the latent code $\bm{z}$. For \textit{X}MNIST datasets, the VAE has an architecture of $512-128-32-32-128-512$ where 512 is the dimension of features extracted in the first step and 32 is the dimension of the latent space. The ReLU layer was employed after each fully-connected layer for non-linearity. For the classifier in step 3, we used a simple two-layer linear neural network (no hidden layer) across all experiments.
We used the Adam optimizer to train the CCVAE with the learning rate of $1e-3$ for a fixed number of epochs (50 epochs for BaggageXray20 and Office-Home, 10 epochs for MNIST datasets). The value of $\lambda$ was dynamically adjusted by a gradual warm-up strategy \cite{goyal2017accurate} from 0 up to 0.2 to facilitate the model training.
\subsection{Experimental Results} 
We compare the performance of CCVAE with three baseline models, two ZSL methods (i.e. BiDiLEL\cite{wang2017zero} and CADA-VAE \cite{schonfeld2019generalized}) adapted for GZSDA and one existing GZSDA method LPP \cite{wang2019unifying}. We do not consider the ZSDA methods in \cite{peng2018zero} and \cite{wang2019conditional} because the former requires paired images and the latter is innately unable to discriminate seen and unseen classes. \textbf{Source Only} uses only source domain data and the 1 Nearest Neighbor (1NN) classifier. The \textbf{Baseline(1NN/NN)} uses training data from both domains (i.e. $\mathcal{D}^{s}$ and $\mathcal{D}^{t}$) and a simple classifier 1NN or Neural Networks (NN) with the same architecture as that used in step 3 of CCVAE. The ZSL methods \textbf{BiDiLEL} and \textbf{CADA-VAE} are used by taking the class means of source samples as the class-level semantic representations.

Following the generalized ZSL works \cite{xian2019f}, we report the mean per-class classification accuracy for seen and unseen classes and their harmony mean ($Acc_{seen}$, $Acc_{unseen}$ and $H$) except for \textit{X}MNIST datasets we only report $H$ due to the limited space. Our experimental results are shown in Tables \ref{table:xray}-\ref{table:digits}. It can be seen from Table \ref{table:xray} that the discrepancy of data distribution between regular and X-ray domains is significant as the \textbf{Source Only} method achieves low accuracy on both adaptation tasks (Table \ref{table:xray}). When labelled target samples from seen classes are employed, the two baseline methods achieve much better performance on seen classes at the sacrifice of accuracy on the unseen classes. The ZSL method BiDiLEL generally performs well on seen classes but poorly on unseen classes due to the notorious issue of overfitting to the seen classes in GZSL. The GZSL method CADA-VAE is able to balance the performance on seen and unseen classes and hence achieves higher values of $H$ but is still outperformed by LPP and our proposed CCVAE due to the fact that the traditional ZSL methods cannot take advantage of the source domain data properly. The proposed CCVAE achieves the highest $H$ values by improving the recognition accuracy of unseen classes whilst maintaining the accuracy of seen classes. The experimental results on the Office-Home dataset (Table \ref{table:officehome}) show similar phenomenon observed on the BaggageXray dataset. CCVAE outperforms other comparative methods on three tasks in terms of $Acc_{unseen}$ and $H$.

In Table \ref{table:digits}, $H$ values are reported for 18 sub-tasks for \textit{X}MNIST datasets. The task of $Color \to Gray$ is an easy one so that even using source data only can achieve as good performance as other more advanced methods except the ZSL methods which, again, suffer from the issue of overfitting to seen classes hence result in low $H$ values. In terms of the other two tasks (i.e. $Gray\to Color$ and $Neg\to Color$) reported in Table \ref{table:digits}, our proposed CCVAE significantly outperforms the comparative methods in 10 out of 12 sub-tasks especially when there are more unseen classes (i.e. EMNIST with 26 classes serving as the unseen dataset).

In summary, our proposed CCVAE is able to handle the GZSDA problem effectively in varying settings across different datasets and outperforms contemporary methods consistently and more significantly in the most challenging scenarios. 

\begin{table*}[!htbp]
    \centering
    {
        \centering
        \caption[]{Experimental results (\%) on BaggageXray dataset with 10 unseen classes (mean and standard error of the mean (SEM) over five different seen/unseen class splits are reported).}
        \label{table:xray}
        \resizebox{0.6\columnwidth}{!}{%
            \begin{tabular}{l| ccc| ccc}
                \hline
                \multirow{2}{*}{Method} & \multicolumn{3}{c|}{Regular $\to$ Xray} &\multicolumn{3}{c}{Xray $\to$ Regular}\\
                \cline{2-7}
                & $Acc_{seen}$ & $Acc_{unseen}$ &  $H$ & $Acc_{seen}$ & $Acc_{unseen}$ &  $H$  \\ \hline
                Source Only&$23.4\pm3.0$ & $20.4\pm3.0$ & $20.2\pm0.7$&$47.9\pm4.0$ & $42.7\pm4.0$ & $43.7\pm1.3$\\
                Baseline(1NN)&$75.0\pm2.4$ & $1.9\pm0.5$ & $3.6\pm0.9$&$93.8\pm1.5$ & $12.6\pm1.3$ & $22.1\pm2.0$\\
                Baseline(NN) &$84.3\pm1.9$&$2.5\pm0.4$&$4.8\pm0.7$&$\bf 95.0\pm0.8$&$20.4\pm3.8$&$32.6\pm5.5$\\
                BiDiLEL \cite{wang2017zero} &$80.8\pm2.2$ & $8.2\pm0.4$ & $14.9\pm0.7$&$94.6\pm0.9$ & $2.8\pm0.6$ & $5.5\pm1.1$\\
                CADA-VAE \cite{schonfeld2019generalized} & $47.3\pm6.7$& $\bf 24.3\pm5.0$& $31.6\pm4.3$& $73.0\pm7.3$& $26.1\pm4.1$& $38.3\pm4.9$\\ 
                LPP \cite{wang2019unifying}&$\bf 85.7\pm1.6$ & $10.2\pm1.1$ & $18.1\pm1.7$&$92.9\pm1.3$ & $30.4\pm2.4$ & $45.6\pm2.9$\\     
                CCVAE&$70.4\pm1.8$&$23.3\pm3.0$&$\bf 34.5\pm3.2$&$87.4\pm0.8$&$\bf 44.5\pm3.5$&$\bf 58.6\pm3.3$\\
                \hline
            \end{tabular}%
    }
}
\end{table*}

\begin{table*}[!htbp]
    \centering
    {
        \centering
        \caption[]{Experimental results (\%) on Office-Home dataset with 30 unseen classes (mean and standard error of the mean (SEM) over five different seen/unseen class splits are reported).}
        \label{table:officehome}
        \resizebox{0.9\columnwidth}{!}{%
            \begin{tabular}{l|ccc|ccc|ccc}
                \hline
                \multirow{2}{*}{Method} & \multicolumn{3}{c|}{ClipArt $\to$ Art} &\multicolumn{3}{c|}{ClipArt $\to$ Product}&\multicolumn{3}{c}{ClipArt $\to$ RealWorld}\\
                \cline{2-10}
                & $Acc_{seen}$ & $Acc_{unseen}$ &  $H$  & $Acc_{seen}$ & $Acc_{unseen}$ &  $H$ & $Acc_{seen}$ & $Acc_{unseen}$ &  $H$  \\ \hline
                Source Only&$48.0\pm0.7$&$44.2\pm0.9$&$46.0\pm0.1$&$55.6\pm0.8$&$57.2\pm0.9$&$56.3\pm0.1$&$59.3\pm1.2$&$59.3\pm1.4$&$59.2\pm0.1$\\
                Baseline(1NN)&$61.0\pm0.2$&$32.3\pm0.9$&$42.2\pm0.8$&$85.3\pm0.5$&$44.6\pm1.1$&$58.6\pm0.9$&$81.0\pm1.6$&$45.3\pm1.1$&$58.0\pm0.7$\\
                Baseline(NN)&$72.6\pm0.4$&$33.0\pm1.3$&$45.3\pm1.1$
                &$88.2\pm0.3$&$50.0\pm1.6$&$63.8\pm1.3$
                &$86.7\pm0.6$&$48.5\pm1.6$&$62.1\pm1.2$\\
                BiDiLEL \cite{wang2017zero} &$\bf 74.7\pm0.8$&$4.5\pm0.4$&$8.4\pm0.8$&$\bf 89.5\pm0.3$&$6.0\pm0.7$&$11.2\pm1.2$&$\bf 87.3\pm0.8$&$5.0\pm0.6$&$9.4\pm1.0$\\
                CADA-VAE \cite{schonfeld2019generalized} & $53.3\pm1.5$& $27.9\pm3.0$& $36.5\pm2.5$& $77.7\pm1.0$& $37.7\pm1.6$& $50.7\pm1.6$& $73.0\pm2.4$& $43.7\pm1.5$& $54.7\pm1.1$\\
                LPP \cite{wang2019unifying} &$72.1\pm0.8$&$48.1\pm0.8$&$57.7\pm0.7$&$87.6\pm0.4$&$58.8\pm1.5$&$70.3\pm1.1$&$86.0\pm0.9$&$59.4\pm2.0$&$70.1\pm1.2$\\  
                CCVAE   &$69.2\pm0.8$&$\bf 51.8\pm0.5$&$\bf 59.2\pm0.3$
                        &$85.4\pm0.5$&$\bf 63.6\pm1.7$&$\bf 72.8\pm1.0$
                        &$83.3\pm0.7$&$\bf 65.1\pm1.8$&$\bf 73.0\pm0.9$\\
                \hline
            \end{tabular}%
        }
    }
\end{table*}

\begin{table*}[!htbp]
    \centering
    {
        \centering
        \caption[]{Experimental results on \textit{X}MNIST datasets (mean and standard deviation of $H$ over five trials are reported).}
        \label{table:digits}
        \resizebox{0.7\columnwidth}{!}{%
            \begin{tabular}{l|l|cccccc}
                \hline
                \multirow{2}{*}{Domains} &\multirow{2}{*}{Method} & \multicolumn{2}{c}{Seen: MNIST} & \multicolumn{2}{c}{Seen: FMNIST} & \multicolumn{2}{c}{Seen: EMNIST}\\
                \cline{3-8}
                & &  FMNIST & EMNIST & MNIST & EMNIST & MNIST & FMNIST\\
                \hline
                \multirow{7}{*}{$Gray \to Color$}& Source Only & $7.3 \pm 0.1$ & $2.5 \pm 0.1$ & $6.1 \pm 0.2$ & $3.5 \pm 0.1$ & $2.2 \pm 0.1$ & $3.3 \pm 0.1$ \\
                & Baseline (1NN)& $46.3 \pm 0.6$ & $30.2 \pm 0.4$ & $38.2 \pm 0.5$ & $17.1 \pm 0.2$ & $51.8 \pm 0.7$ & $47.6 \pm 0.4$ \\
                & Baseline (NN) & $50.0\pm0.1$& $36.2\pm0.1$& $47.9\pm0.1$& $27.5\pm0.1$& $42.7\pm0.1$& $42.5\pm0.1$\\
                & BiDiLEL \cite{wang2017zero} & $39.9 \pm 2.0$ & $35.8 \pm 1.1$ & $30.6 \pm 1.9$ & $13.9 \pm 1.4$ & $52.8 \pm 1.9$ & $37.6 \pm 1.6$ \\
                & CADA-VAE \cite{schonfeld2019generalized} & $39.2\pm1.9$& $35.5\pm1.3$& $30.0\pm3.5$& $19.3\pm1.6$& $46.3\pm1.1$& $43.3\pm0.6$\\ 
                & LPP \cite{wang2019unifying} & $61.3 \pm 0.5$ & $43.5 \pm 0.6$ & $58.8 \pm 0.4$ & $39.1 \pm 0.4$ & $68.8 \pm 0.3$ & $\bf 58.8 \pm 0.2$ \\  
                & CCVAE& $\bf 63.9\pm0.2$& $\bf 61.4\pm0.0$& $\bf 68.1\pm1.1$& $\bf 45.4\pm0.3$& $\bf 71.8\pm0.4$& $53.8\pm2.0$\\
                \hline
                \multirow{7}{*}{$Color \to Gray$}& Source Only & $86.5 \pm 0.6$ & $89.0 \pm 0.3$ & $87.1 \pm 0.6$ & $81.4 \pm 0.6$ & $89.6 \pm 0.3$ & $80.8 \pm 0.3$ \\
                & Baseline (1NN)& $85.6 \pm 0.5$ & $87.6 \pm 0.3$ & $90.9 \pm 0.2$ & $85.2 \pm 0.2$ & $90.7 \pm 0.1$ & $82.8 \pm 0.5$ \\
                & Baseline (NN) & $\bf 89.6\pm0.0$& $89.1\pm0.0$& $\bf 92.5\pm0.0$& $\bf 87.8\pm0.0$& $91.0\pm0.0$& $\bf 86.9\pm0.0$\\
                & BiDiLEL \cite{wang2017zero} & $29.0 \pm 2.6$ & $31.8 \pm 2.1$ & $18.7 \pm 2.9$ & $10.1 \pm 3.4$ & $53.7 \pm 4.3$ & $38.9 \pm 2.7$ \\
                & CADA-VAE \cite{schonfeld2019generalized} & $39.1\pm2.2$& $38.9\pm2.7$& $47.3\pm4.5$& $31.2\pm5.1$& $52.9\pm2.0$& $45.8\pm4.4$\\ 
                & LPP \cite{wang2019unifying} & $86.6 \pm 0.2$ & $80.3 \pm 0.2$ & $90.9 \pm 0.1$ & $81.3 \pm 0.3$ & $85.2 \pm 0.4$ & $81.5 \pm 0.3$ \\
                & CCVAE& $88.6\pm0.1$& $\bf 90.2\pm0.1$& $92.0\pm0.0$& $87.1\pm0.0$& $\bf 91.9\pm0.0$& $85.8\pm0.1$\\
                
                \hline
                \multirow{7}{*}{$Neg. \to Color$}& Source Only& $5.1 \pm 0.3$ & $1.9 \pm 0.1$ & $5.0 \pm 0.4$ & $1.4 \pm 0.0$ & $1.8 \pm 0.1$ & $2.7 \pm 0.0$  \\
                & Baseline (1NN) & $41.4 \pm 0.9$ & $28.8 \pm 0.6$ & $25.5 \pm 0.5$ & $9.6 \pm 0.1$ & $64.9 \pm 0.4$ & $46.9 \pm 1.1$ \\
                & Baseline (NN)& $49.7\pm0.1$& $36.7\pm0.2$& $44.5\pm0.1$& $23.9\pm0.2$& $55.9\pm0.0$& $40.8\pm0.1$\\
                & BiDiLEL \cite{wang2017zero} & $36.6 \pm 2.6$ & $31.3 \pm 1.1$ & $27.0 \pm 2.1$ & $13.5 \pm 0.8$ & $50.9 \pm 1.7$ & $36.6 \pm 1.7$  \\
                &CADA-VAE \cite{schonfeld2019generalized}& $39.3\pm1.7$& $34.5\pm1.8$& $26.7\pm1.4$& $18.4\pm4.6$& $42.7\pm1.1$& $41.0\pm0.9$\\
                & LPP \cite{wang2019unifying} & $68.1 \pm 0.6$ & $45.2 \pm 0.6$ & $54.1 \pm 0.8$ & $30.5 \pm 0.4$ & $69.2 \pm 0.2$ & $\bf 66.7 \pm 0.2$ \\
                & CCVAE & $\bf 70.6\pm0.4$& $\bf 63.2\pm0.5$& $\bf 68.4\pm1.1$& $\bf 46.9\pm0.9$& $\bf 72.4\pm1.0$& $62.0\pm1.4$\\
                \hline
            \end{tabular}%
        }
    }
\end{table*}

\section{Discussion and Conclusion}\label{sec:conclusion}
The key to GZSDA is to overcome the overfitting to seen classes so that ZSDA methods such as \cite{wang2019conditional} and generic generative models such as CycleGAN \cite{zhu2017unpaired} do not apply. LPP achieves this goal by mapping the source and target data into a common subspace of lower dimension with a unified linear projection. However, it only works well when the domain shift is not significant. Moreover, the solution to LPP involves the eigen-decomposition of a matrix relating to the number of training samples hence is not scalable.

The generative models (e.g., CADA-VAE and CCVAE) address the overfitting issue by generating synthetic data for the unseen classes. Our proposed CCVAE was inspired by CADA-VAE and has a similar framework but is essentially different from it. Firstly, CADA-VAE generates features from class-wise attribute vectors, restricting the intra-class variations of the synthetic features whilst CCVAE generates features from individual samples. Secondly, CADA-VAE employs domain-specific VAE for source and target domains whilst CCVAE uses a unified VAE to promote the preserving of class information in the latent space. As a result, to generate both domain and class discriminative features, the generative model in CADA-VAE is conditioned on class information whilst CCVAE is conditioned on domain information. Finally, CCVAE is used to generate not only target domain features but also source domain features to augment the real training data and a unified classifier is trained for both domains in CCVAE (Step 3).

In conclusion, our proposed CCVAE is an effective approach to the GZSDA problems. In addition, our proposed BaggageXray dataset provides a challenging test bed for future researches in GZSDA as well as other domain adaptation problems given the fact it arises from a real-world application in aviation security screening and the unique spectral X-ray imagery.

\section{Broader Impacts}
This work addresses a problem arising from real-world applications. The traditional ZSL problem setting has its intrinsic limitation in that the class-level semantic representations are insufficient to model the intra-class variability. In contrast, GZSDA provides a possibility of recognizing novel classes without the need of training data for these classes in the target domain. For example, in the X-ray based aviation security screening application, the definition of threat objects evolves rapidly since the terrorists never stop devising new weapons/explosive-devices. Automatic threat recognition algorithms used in the checkpoint can be updated conveniently without collecting X-ray data of new threat items which is usually the most time-consuming procedure \cite{wang2019approach}.  

The other potential application is cross-scene (e.g., day and night, rainy and foggy) adaptation of perception algorithms in autonomous driving. One issue in the object detection for autonomous driving is the long-tail distribution problem. It is even difficult to handle this issue in adverse weather conditions  \cite{bijelic2018benchmarking} where GZSDA can be employed to transfer knowledge learned from daytime/sunny domain to night/rainy/foggy domain so as to avoid collecting the same amount of data for training.

\bibliography{ref}
\bibliographystyle{plain}

\end{document}